\documentclass{article}

\usepackage{PRIMEarxiv}

\usepackage[utf8]{inputenc}
\usepackage[T1]{fontenc}
\usepackage{microtype}

\usepackage{amsmath,graphicx,hyperref}
\usepackage{amssymb,amsfonts,bm,mathtools}
\usepackage{siunitx}
\usepackage{dsfont}
\usepackage{booktabs}
\usepackage{multirow}
\usepackage{float}
\usepackage{subcaption}
\usepackage{placeins}
\usepackage{algorithm}
\usepackage{algorithmic}
\usepackage{xcolor}
\usepackage{enumitem}
\setlist{nosep,leftmargin=2em}

\newcommand{\ninept}{}

\title{UNIFIED MULTIMODAL COHERENT FIELD: \\SYNCHRONOUS SEMANTIC-SPATIAL-VISION FUSION FOR BRAIN TUMOR SEGMENTATION}

\author{
Mingda Zhang \and Yuyang Zheng \and Ruixiang Tang \and Jingru Qiu \and Haiyan Ding$^{*}$\\
School of Software, Yunnan University, Kunming 650500, China\\
\texttt{\{zhangmingda, zyy0906,tangruixiang, choujingru\}@stu.ynu.edu.cn, dinghaiyan@ynu.edu.cn}
}

\begin{document}
\ninept
\maketitle

\begin{abstract}

Brain tumor segmentation requires accurate identification of hierarchical regions including whole tumor (WT), tumor core (TC), and enhancing tumor (ET) from multi-sequence Magnetic Resonance Imaging (MRI) images. Due to tumor tissue heterogeneity, ambiguous boundaries, and contrast variations across MRI sequences, methods relying solely on visual information or post-hoc loss constraints show unstable performance in boundary delineation and hierarchy preservation. To address this challenge, we propose the Unified Multimodal Coherent Field (UMCF) method. This method achieves synchronous interactive fusion of visual, semantic, and spatial information within a unified 3D latent space, adaptively adjusting modal contributions through parameter-free uncertainty gating, with medical prior knowledge directly participating in attention computation, avoiding the traditional “process-then-concatenate” separated architecture. On Brain Tumor Segmentation (BraTS) 2020 and 2021 datasets, UMCF+nnU-Net achieves average Dice coefficients of 0.8579 and 0.8977 respectively, with an average 4.18\% improvement across mainstream architectures. By deeply integrating clinical knowledge with imaging features, UMCF provides a new technical pathway for multimodal information fusion in precision medicine.

\end{abstract}

% Replace spconf keywords env with PRIMEarxiv \keywords (content preserved)

\section{INTRODUCTION}
\label{sec:intro}

Brain tumor segmentation is one of the most challenging fundamental tasks in neuro-oncology. Its results directly impact clinical decisions including preoperative assessment, treatment monitoring, and radiotherapy planning. While multi-sequence MRI (including T1-weighted (T1), T1-weighted contrast-enhanced (T1ce), T2-weighted (T2), and Fluid-Attenuated Inversion Recovery (FLAIR)) provides complementary contrast information, actual lesions typically exhibit tissue heterogeneity, irregular morphology, and ambiguous boundaries. Additionally, the nested hierarchical relationship between ET, TC, and WT (ET$\subset$TC$\subset$WT) requires global consistency from models\cite{BraTS2018Tasks}. These factors collectively cause solutions relying on single-modal information or post-processing corrections to compromise on boundary delineation and hierarchy preservation\cite{Litjens2017Survey}.

Current research explores two main directions: designing powerful visual feature extraction networks and improving multimodal fusion strategies. However, Convolutional Neural Networks (CNNs) struggle with global feature relationships\cite{Ronneberger2015UNet,Cicek2016_3DUNet}, while Transformers incur high computational costs\cite{ViT2020,TransUNet2021}. Most fusion approaches simply stack multimodal images with equal weights, ignoring MRI sequences' differential sensitivity to specific pathological regions\cite{BraTS2018Tasks}. Recent attention-based methods still perform multimodal interaction after rather than during feature extraction, failing to leverage modality-specific associations (T1/T1ce for tumor core, FLAIR/T2 for edema) in attention computation\cite{BiomedCLIP2023}.

This paper's main contributions are:
\begin{enumerate}
\item We propose the UMCF framework for synchronous fusion of visual, semantic, and spatial information within a unified 3D latent space, embedding medical priors directly into attention computation.

\item We design a parameter-free coordination mechanism including Zero-parameter Semantic-Spatial Channel Modulation (ZSCM), Parameter-Free Uncertainty Gating (PFUG), and convex optimization updates, improving cross-modal consistency without additional trainable parameters.
\end{enumerate}

\section{METHOD}
\label{sec:method}

\subsection{Overall Architecture}

\begin{figure}[t!]
  \centering
  \centerline{\includegraphics[width=17cm]{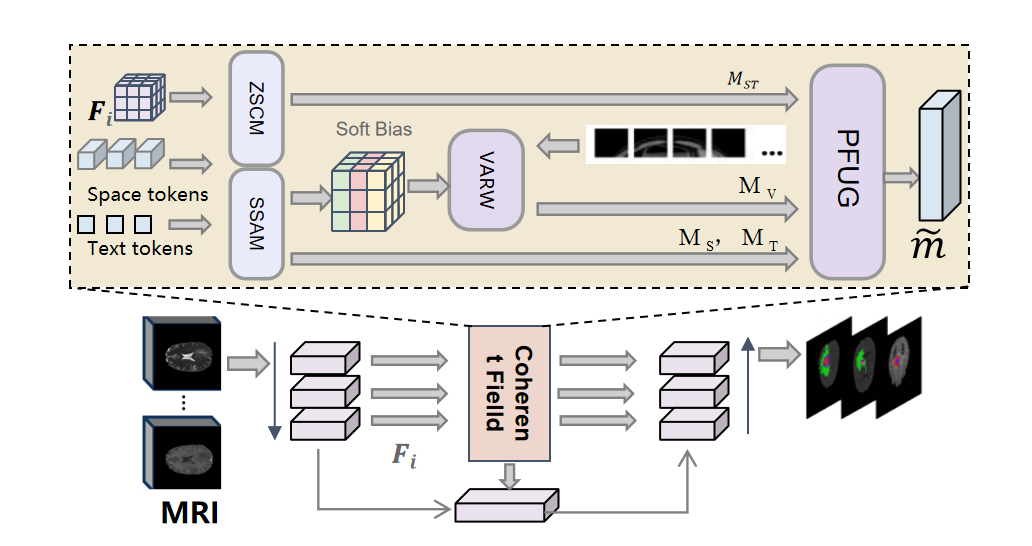}}
  \caption{UMCF overall architecture diagram}
  \label{fig:overview}
\end{figure}

UMCF changes the traditional ``process-then-concatenate'' pattern by constructing a unified 3D latent space where visual (V), semantic/text (T), and spatial prior (S) information interact in real-time. This plug-and-play fusion layer, inserted between encoder and decoder of U-Net architectures\cite{Ronneberger2015UNet,Cicek2016_3DUNet}, receives multi-sequence MRI images and clinical text descriptions as input. As shown in Figure~\ref{fig:overview}, multi-sequence MRI images pass through an encoder to extract multi-scale features $F_i$. Multimodal fusion occurs at the bottleneck layer through coordinated processing by sub-modules: ZSCM, Semantic-Spatial Attention Modulation(SSAM), Visual Attention Read-Write with Medical Priors(VARW), and PFUG, producing segmentation results that satisfy medical hierarchical relationships.

\subsection{Data Preprocessing and Feature Extraction}

Four MRI sequences (T1, T1ce, T2, FLAIR) undergo registration, intensity normalization, and resampling to form $X\!\in\!\mathbb{R}^{H\times W\times D\times C}$. Clinical text is parsed using medical NLP tools and encoded via BiomedCLIP or ClinicalBERT\cite{BiomedCLIP2023,Alsentzer2019ClinicalBERT}. A U-Net encoder produces multi-scale feature pyramid $\{E_s\}_{s=1}^{L}$. The bottleneck feature $B\!=\!E_L$ is projected to $d$ dimensions through $1{\times}1{\times}1$ convolution, forming initial latent field $F^{(0)}\!\in\!\mathbb{R}^{H_b\times W_b\times D_b\times d}$. All feature vectors are L2-normalized to the unit sphere, using cosine similarity $\mathrm{sim}(a,b)=a^\top b$ with temperature parameter $\tau>0$.

\subsection{Multimodal Information Encoding and Coordinated Fusion Mechanism}

Visual Token Construction.
Visual tokens aggregate features from local image regions:
\begin{equation}
\label{eq:vis-token}
V_i=\frac{\frac{1}{|P_i|}\sum_{x\in P_i}E(x)}{\left\|\frac{1}{|P_i|}\sum_{x\in P_i}E(x)\right\|_2}\,,
\end{equation}
where $P_i$ indexes the $i$-th cubic block (e.g., $8{\times}8{\times}8$ voxels) at bottleneck resolution. Average pooling within each block provides noise-robust region representations, while normalization ensures all visual tokens lie on the unit sphere for consistent similarity computation.

Semantic Token Construction.
Medical concepts from clinical text are converted to semantic vectors aligned with visual features:
\begin{equation}
\label{eq:sem-token}
T_j=\frac{\frac{1}{|\mathcal P_j|}\sum_{w\in\mathcal P_j}e(w)}{\left\|\frac{1}{|\mathcal P_j|}\sum_{w\in\mathcal P_j}e(w)\right\|_2}\,,
\end{equation}
where each medical phrase $\mathcal P_j$ (e.g., ``ring enhancement'', ``central necrosis'') obtains word embeddings $e(w)$ through pre-trained medical text encoders\cite{BiomedCLIP2023,Alsentzer2019ClinicalBERT}. Multiple word vectors within a phrase are merged via average pooling then normalized. The semantic prototype $\bar T$, computed as the equal-weight average of all semantic tokens, represents the overall semantic features of the current case.

Spatial Token Construction.
Spatial prior information captures tumor position, morphology, and topology from current segmentation probability maps $P_c(\cdot)$ where $c\!\in\!\{\mathrm{ET,TC,WT}\}$:
\begin{equation}
\label{eq:center}
\mu_c=\frac{\sum_{x\in\Omega} P_c(x)\,x}{\sum_{x\in\Omega} P_c(x)}\,,
\end{equation}
\begin{equation}
\label{eq:covariance}
\Sigma_c=\frac{\sum_{x\in\Omega} P_c(x)\,(x-\mu_c)(x-\mu_c)^\top}{\sum_{x\in\Omega} P_c(x)}\,,
\end{equation}
\begin{equation}
\label{eq:eigs}
(\lambda_1^c\!\ge\!\lambda_2^c\!\ge\!\lambda_3^c)=\operatorname{eig}(\Sigma_c),
\end{equation}
\begin{equation}
\label{eq:sdt}
\overline{D}_c=\frac{1}{|\Omega|}\sum_{x\in\Omega} \mathrm{SDT}_c(x)\!,
\end{equation}
\begin{equation}
\label{eq:Shier}
S^{\mathrm{hier}}_c=\frac{\big[\mu_c^\top,\ \lambda_1^c,\lambda_2^c,\lambda_3^c,\ \overline{D}_c\big]^\top}
{\left\|\big[\mu_c^\top,\ \lambda_1^c,\lambda_2^c,\lambda_3^c,\ \overline{D}_c\big]^\top\right\|_2}\,.
\end{equation}
These statistics jointly describe tumor spatial characteristics: centroid $\mu_c$ indicates tumor position, eigenvalues $\lambda_{1,2,3}^c$ of covariance matrix $\Sigma_c$ reflect tumor extension along principal directions, and the average Signed Distance Transform (SDT) $\overline{D}_c$---which measures each voxel's signed distance to the nearest boundary---characterizes boundary thickness and inside-outside relationships\cite{Maurer2003EDT,Felzenszwalb2012DT}. After concatenating and normalizing these features, we obtain hierarchical structure token $S^{\mathrm{hier}}_c$. Additional topological features (neighborhood smoothness, boundary gradients, surface-to-volume ratio) form the complete spatial token set $\{S_k\}$, with spatial prototype $\bar S$ as their average.

With token representations established, we construct their interaction mechanism. First, semantic information requires spatial ``grounding'' through the semantic field:
\begin{equation}
\label{eq:phiT}
\phi_T(x)=\sigma\!\left(\frac{\mathrm{sim}\big(F(x),\ \bar T\big)}{\tau}\right)\,.
\end{equation}
This semantic field $\phi_T$ acts as a soft spatial attention map, evaluating consistency between each voxel position and overall semantics. High-response regions indicate strong alignment between visual features and clinical descriptions, guiding subsequent attention mechanisms.

% Figure 2 moved to double-column format
\begin{figure*}[t!]
  \centering
  \includegraphics[width=17cm]{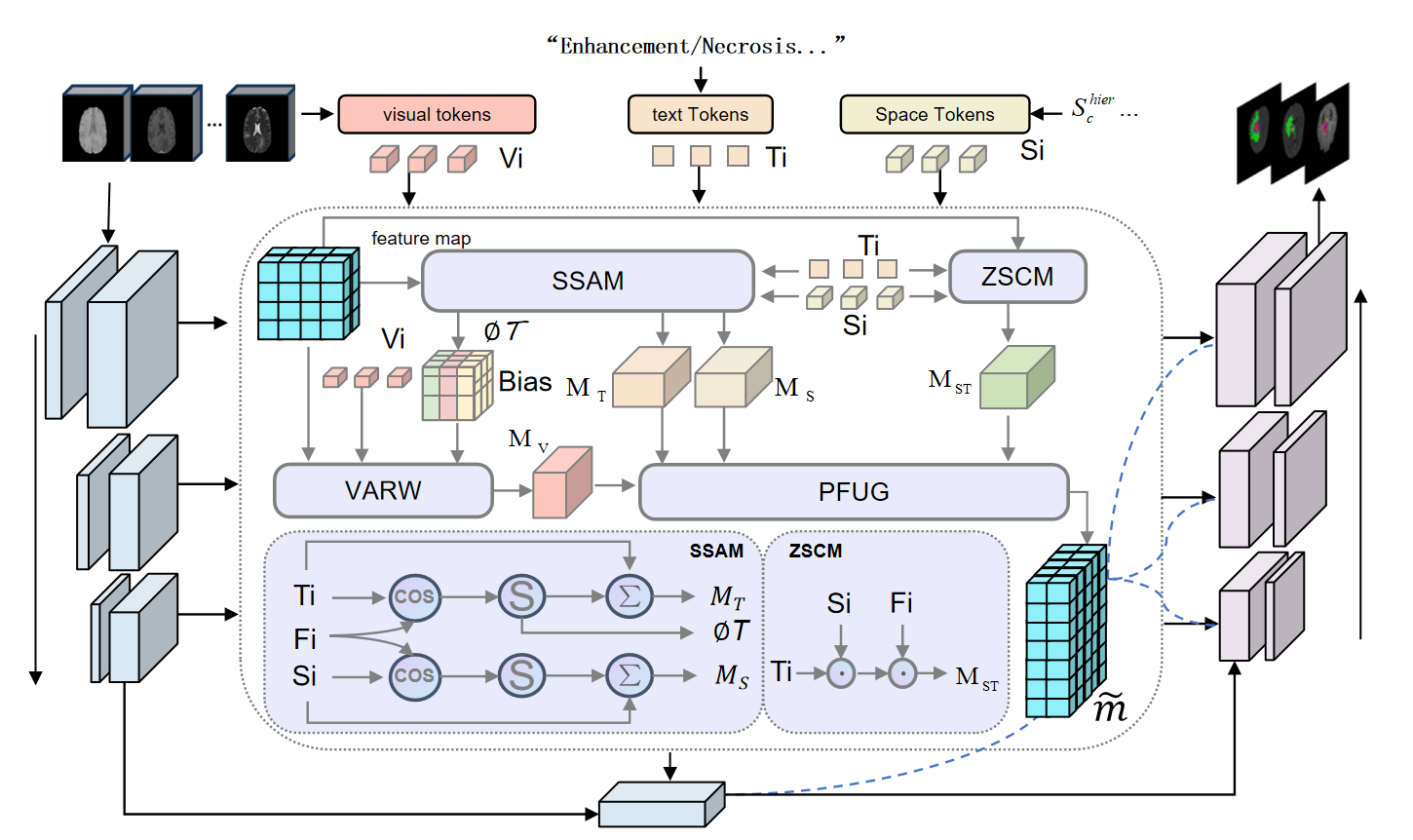}
  \caption{UMCF core module detail diagram. Shows the semantic-spatial collaboration mechanism of ZSCM/SSAM, visual attention read-write of VARW, and uncertainty-weighted fusion process of PFUG. Visual tokens (Vi), spatial tokens (Si), and text tokens (Ti) participate in attention calculation through soft bias mechanism, producing multi-path messages that ultimately fuse into voxel consensus ($\tilde{m}$).}
  \label{fig:detail}
\end{figure*}

Based on this semantic field foundation, UMCF achieves deep multimodal fusion through four coordinated modules (Figure~\ref{fig:detail}). These modules transform tokens from three modalities into four complementary message streams ($m_S$, $m_T$, $m_V$, $m_{ST}$), ultimately fused into unified voxel representation $\tilde{m}$.

Visual Attention Read-Write with Medical Priors (VARW).
Attention incorporates semantic and spatial biases:
\begin{gather}
\label{eq:alphaV}
\alpha^V_{x,i}=\frac{\exp\!\left(\big[\mathrm{sim}(F(x),V_i)+\mu_V(x,i)\big]/\tau\right)}
{\sum\limits_{p}\exp\!\left(\big[\mathrm{sim}(F(x),V_p)+\mu_V(x,p)\big]/\tau\right)}\!,\\
\label{eq:mV}
m_V(x)=\sum_{i}\alpha^V_{x,i}\,V_i,\\
\label{eq:muV}
\mu_V(x,i)=\log\!\big(1+\phi_T(x)\big)\ -\ r_{\mathrm{hier}}(x)\ -\ r_{\mathrm{topo}}(x)\,.
\end{gather}
The bias $\mu_V$ includes: semantic encouragement $\log(1+\phi_T)$ with logarithmic scaling, hierarchical penalty $r_{\mathrm{hier}}$ for ET$\subset$TC$\subset$WT violations\cite{BraTS2018Tasks}, and topological penalty $r_{\mathrm{topo}}$ for discontinuous boundaries. Thus $m_V(x)$ represents medically-constrained visual evidence.

Semantic-Spatial Attention Modulation (SSAM).
Modality-specific messages aggregate relevant tokens:
\begin{equation}
\label{eq:mq}
m_{q}(x)=\sum_{z\in\mathcal{I}_{q}}
\operatorname{softmax}_{z}\!\left(\frac{\mathrm{sim}\big(F(x),\,z\big)}{\tau}\right)\,z\,,\ q\in\{T,S\}\,.
\end{equation}
Semantic messages ($q\!=\!T$) soft-select relevant medical concepts, while spatial messages ($q\!=\!S$) combine position, scale, and boundary information based on feature similarity.

Zero-parameter Semantic-Spatial Channel Modulation (ZSCM).
Cross-modal synergy through element-wise multiplication:
\begin{equation}
\label{eq:mTS}
m_{TS}(x)=\big(\bar T\odot \bar S\big)\odot F(x)\,.
\end{equation}
Dimensions activated by both semantic and spatial prototypes are enhanced, while conflicting responses are suppressed, filtering features endorsed by multiple modalities without trainable parameters.

Parameter-Free Uncertainty Gating (PFUG).
Four message streams are adaptively fused based on reliability:
\begin{equation}
\label{eq:gating}
\tilde m(x)=\sum_{q\in\{V,T,S,TS\}}
\frac{\exp\!\big(-u_q(x)\big)}
{\sum\limits_{p\in\{V,T,S,TS\}}\exp\!\big(-u_p(x)\big)}\,m_q(x)\,.
\end{equation}
Uncertainty measures: $u_V$ uses prediction entropy (ambiguity indicator), $u_T$ measures text-vision inconsistency, $u_S$ evaluates continuity residuals, and $u_{TS}$ averages semantic-spatial uncertainties. Normalized to $[0,1]$, these weights allow relying on semantic-spatial priors in ambiguous regions while prioritizing visual evidence in clear regions.

\subsection{Synchronous Convex Optimization Update}

The fused message $\tilde m(x)$ integrates with current field representation via:
\begin{equation}
\label{eq:update}
F^{(t+1)}(x)=(1-\lambda)\,F^{(t)}(x)+\lambda\,\tilde m(x)\,.
\end{equation}
With $\lambda\!\in\!(0,1)$, this convex combination ensures convergence without oscillation. After 2-4 iterations with channel-wise renormalization, the converged field $F^\star$ passes to the decoder, producing full-resolution segmentation maps $P$ through layer-wise upsampling.

\section{EXPERIMENTS AND RESULTS}

\subsection{Datasets and Experimental Setup}

We evaluate UMCF on BraTS 2020 (369 training, 125 validation) and BraTS 2021 (1251 training, 219 validation) benchmark datasets\cite{BraTS2020Stats,BraTS2021Stats}. Each case includes four co-registered MRI sequences (T1, T1ce, T2, FLAIR) at $1{\times}1{\times}1$mm³ resolution with expert annotations for three nested hierarchical structures: enhancing tumor (ET), tumor core (TC), and whole tumor (WT), satisfying ET$\subset$TC$\subset$WT\cite{BraTS2018Tasks}.

Experiments utilize NVIDIA A100 GPUs with Dice coefficients as the primary metric. The loss combines soft Dice and weighted cross-entropy\cite{VNet2016Dice,CEplusDice2022}. UMCF integrates as a plug-and-play module into 3D U-Net and nnU-Net architectures\cite{Cicek2016_3DUNet,Isensee2021nnUNet} with base channels 32, latent dimension $d\!=\!256$, using AdamW optimizer with OneCycleLR scheduling\cite{AdamW2019,OneCycle2017}. Data augmentation includes rotation (±15°), flipping (50\%), Gamma correction (0.8-1.2), and modality-specific intensity enhancement. Note that due to differing calculation standards for HD95 distance in different literature (e.g., using surface-distance library or MedPy library) \cite{SurfaceDistanceGitHub,MedPyHD95Issue}, to ensure fairness, this paper does not adopt HD95 metric in method comparisons.

\subsection{Comparison Experiments}

To evaluate UMCF's performance improvement and verify its generalization ability across different data scales and years, we design comprehensive comparative experiments. We select representative methods from 2020 to 2025, covering BraTS competition winning solutions, recently proposed innovative architectures, and methods introducing multimodal information. Table \ref{tab:comparison} shows performance comparisons of various methods on BraTS 2020 and 2021 validation set.

\begin{table*}[t!]
\centering
\linespread{1}\selectfont
\caption{Dice coefficients comparison of different methods on BraTS 2020 and BraTS 2021 datasets}
\label{tab:comparison}
\footnotesize
\setlength{\tabcolsep}{2.8pt}
\begin{tabular}{p{4cm}cccccp{4cm}ccccc}
\toprule
\multicolumn{6}{c}{BraTS 2020} & \multicolumn{6}{c}{BraTS 2021} \\
\cmidrule(r){1-6} \cmidrule(l){7-12}
Method & Year & Avg & WT & TC & ET & Method & Year & Avg & WT & TC & ET \\
\midrule
UMCF + nnU-Net (ours) & 2025 & \textbf{0.8579} & \textbf{0.9110} & 0.8668 & 0.7958 & UMCF + nnU-Net (ours) & 2025 & \textbf{0.8977} & 0.9289 & 0.9066 & 0.8577 \\
CLIP-UNet \cite{zhang2025clipunet} & 2025 & 0.8567 & 0.8994 & \textbf{0.8709} & 0.8005 & Two-branch SR-Net \cite{jia2023twobranch} & 2025 & 0.8970 & 0.9105 & 0.8930 & \textbf{0.8861} \\
nnU-Net (Winner) \cite{Isensee2021nnUNet} & 2020 & 0.8535 & 0.8955 & 0.8506 & \textbf{0.8203} & DeepSeg Ensemble (Winner) \cite{baid2021brats21} & 2021 & 0.8960 & \textbf{0.9294} & 0.8788 & 0.8803 \\
UMCF + 3D U-Net (ours) & 2025 & 0.8505 & 0.9048 & 0.8621 & 0.7846 & SegResNet \cite{cardoso2022monai} & 2025 & 0.8910 & 0.9170 & 0.8960 & 0.8610 \\
FCFDiff-Net \cite{pang2025fcfdiff} & 2025 & 0.8380 & 0.8980 & 0.8300 & 0.7860 & UMCF + 3D U-Net (ours) & 2025 & 0.8874 & 0.9012 & 0.9121 & 0.8488 \\
BU-Net-ASPP \cite{yousef2023bridged} & 2023 & 0.8344 & 0.9073 & 0.8159 & 0.7800 & BU-Net-ASPP-EVO \cite{yousef2023bridged} & 2023 & 0.8740 & 0.9187 & 0.8594 & 0.8434 \\
Modified U-Net \cite{yousef2023unet} & 2023 & 0.8310 & 0.9050 & 0.8070 & 0.7810 & 3D ResUNet \cite{pei2022multimodal} & 2022 & 0.8630 & 0.8190 & \textbf{0.9196} & 0.8503 \\
LATUP-Net \cite{latup2025} & 2024 & 0.8197 & 0.8841 & 0.8382 & 0.7367 & RAL-Net \cite{peiris2022reciprocal} & 2022 & 0.8584 & 0.8138 & 0.9076 & 0.8538 \\
nnU-Net (baseline, ours) & 2025 & 0.8175 & 0.8784 & 0.8498 & 0.7243 & nnU-Net (baseline, ours) & 2025 & 0.8522 & 0.9096 & 0.8477 & 0.7993 \\
3D U-Net (baseline, ours) & 2025 & 0.8088 & 0.8887 & 0.8099 & 0.7277 & 3D U-Net (baseline, ours) & 2025 & 0.8477 & 0.8601 & 0.8716 & 0.8112 \\
\bottomrule
\end{tabular}
\end{table*}

The results demonstrate UMCF's architecture-agnostic nature, with consistent ~4\% performance improvements when integrated with both nnU-Net and simpler 3D U-Net architectures, validating its effectiveness as a plug-and-play module that enhances different backbone networks without architecture-specific modifications. Notably, the most significant gains occur in smaller, ambiguous regions, with TC improving by 6.95\% and ET by 7.38\%, confirming that multimodal fusion particularly benefits difficult cases where visual information alone is insufficient. Furthermore, while CLIP-UNet also incorporates text information for guidance, UMCF achieves superior performance by embedding semantic bias directly into the attention computation process, enabling real-time multimodal interaction throughout the network rather than relying on post-hoc feature concatenation, thus achieving deeper semantic-visual synergy throughout the segmentation process.

\subsection{Ablation Study}
To understand the specific contributions of each component in the UMCF framework and validate the rationality of design decisions, we conduct systematic ablation studies. Experiments observe performance changes by progressively removing or replacing modules, including four-path message passing mechanism, parameter-free uncertainty gating, and synchronous fusion strategy. Table \ref{tab:ablation} shows the performance of various configurations.

% ---- Table 2: same font size as body, a bit smaller overall, wider spacing ----
\begin{table}[t]
\centering
\caption{Ablation study results for UMCF components (based on nnU-Net backbone)}
\label{tab:ablation}

\begin{minipage}{0.92\linewidth} % slightly smaller than text width, no font scaling
\centering
\setlength{\tabcolsep}{5.5pt}    % wider column padding (default ≈6pt; earlier we used ~2pt)
\renewcommand{\arraystretch}{1.15} % a bit more row spacing

\sisetup{
  table-number-alignment = center,
  table-format = 1.4,
  detect-weight = true,
  detect-inline-weight = math
}

\begin{tabular}{
  l
  *{4}{S}
  @{\hskip 12pt}           % extra gap between 2020 and 2021 blocks
  *{4}{S}
}
\toprule
& \multicolumn{4}{c}{BraTS 2020} & \multicolumn{4}{c}{BraTS 2021} \\
\cmidrule(lr){2-5}\cmidrule(lr){6-9}
{Configuration} & {Avg} & {WT} & {TC} & {ET} & {Avg} & {WT} & {TC} & {ET} \\
\midrule
UMCF                 & {\bfseries 0.8579} & {\bfseries 0.9110} & {\bfseries 0.8668} & {\bfseries 0.7958} & {\bfseries 0.8977} & {\bfseries 0.9289} & {\bfseries 0.9066} & {\bfseries 0.8577} \\
w/o $m_V$            & 0.8377 & 0.8989 & 0.8422 & 0.7721 & 0.8835 & 0.9166 & 0.8861 & 0.8478 \\
w/o $m_T$            & 0.8421 & 0.9001 & 0.8473 & 0.7789 & 0.8762 & 0.8979 & 0.8842 & 0.8464 \\
w/o $m_{ST}$         & 0.8395 & 0.8986 & 0.8418 & 0.7780 & 0.8692 & 0.8943 & 0.8789 & 0.8344 \\
w/o $m_S$            & 0.8372 & 0.8987 & 0.8416 & 0.7713 & 0.8646 & 0.9125 & 0.8652 & 0.8161 \\
w/o PFUG             & 0.8317 & 0.8977 & 0.8422 & 0.7551 & 0.8638 & 0.8990 & 0.8753 & 0.8171 \\
Pairwise fusion      & 0.8320 & 0.8921 & 0.8430 & 0.7609 & 0.8570 & 0.8761 & 0.8725 & 0.8224 \\
Baseline             & 0.8175 & 0.8784 & 0.8498 & 0.7243 & 0.8522 & 0.9096 & 0.8477 & 0.7993 \\
\bottomrule
\end{tabular}
\end{minipage}
\end{table}

To analyze component contributions, we conduct ablation studies using nnU-Net backbone. Table~\ref{tab:ablation} presents the quantitative impact of removing individual modules. Results reveal critical insights: removing spatial message $m_S$ causes maximum degradation (average 3.31\% decrease on BraTS 2021, with ET decreasing 4.85\%), confirming its essential role in capturing tumor morphology and maintaining hierarchical relationships\cite{BraTS2018Tasks}. Semantic message $m_T$ contributes 2.15\%, primarily benefiting TC/ET regions where clinical descriptions provide valuable guidance. The parameter-free $m_{ST}$ achieves 2.85\% improvement through channel-level coordination. PFUG enables 3.4\% gain through uncertainty-adaptive fusion. Most critically, replacing synchronous fusion with pairwise fusion—where modalities are combined two at a time rather than simultaneously yields, demonstrating that simultaneous multimodal interaction within a unified latent space, rather than sequential pairwise processing, is fundamental to UMCF's success.
\subsection{Visualization Analysis}

\begin{figure}[htbp]
\centering
\centerline{\includegraphics[width=0.8\columnwidth]{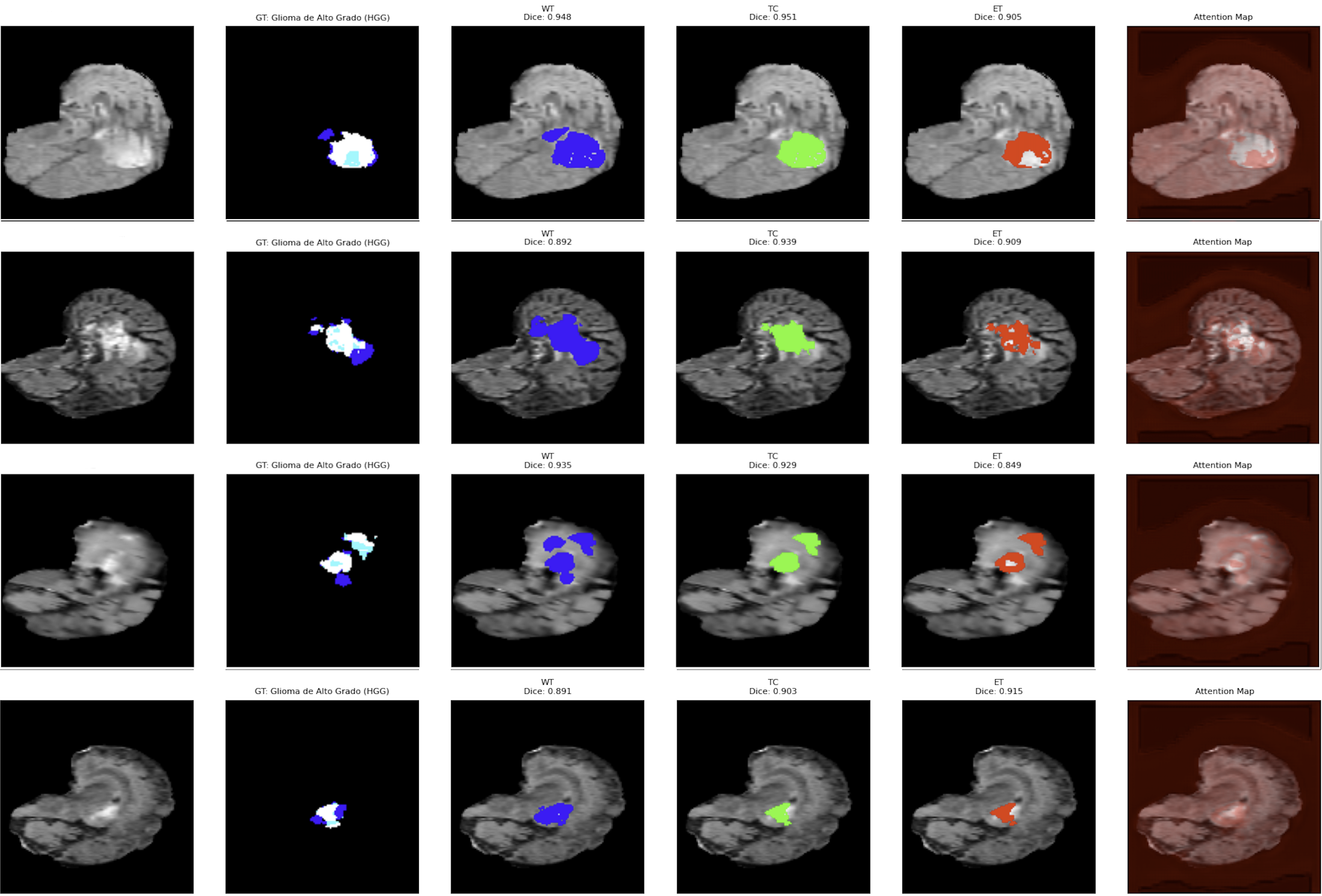}}
\caption{UMCF segmentation visualization results}
\label{fig:visualization}
\end{figure}

Figure~\ref{fig:visualization} illustrates UMCF's segmentation on four representative cases, where blue, green, and orange represent WT, TC, and ET regions respectively. Results demonstrate accurate morphology reconstruction: WT regions (blue) completely envelope lesions without over-segmentation; TC regions (green) precisely capture irregular tumor cores; ET regions (orange), though small and scattered, are correctly identified. Notably, the segmentation boundaries present natural curved morphology without blocky artifacts or discontinuous breaks, with smooth transitions between the three sub-regions while maintaining proper nested relationships (ET$\subset$TC$\subset$WT). This confirms that the collaborative action of semantic field $\phi_T$ and medical bias $\mu_V$ successfully achieves spatial modulation of visual attention through medical prior knowledge, enabling UMCF to produce accurate and consistent segmentations across tumors of varying sizes and morphologies.

\section{CONCLUSION}

UMCF achieves synchronous fusion of visual, semantic, and spatial information within a unified 3D latent space for brain tumor segmentation. Its core innovations include semantic field-guided localization, medical knowledge-constrained attention, parameter-free coordination, and uncertainty-adaptive fusion. Experimental results show significant boundary quality improvement. As a plug-and-play module, UMCF provides an effective solution for multimodal medical image analysis.

% References
\bibliographystyle{IEEEbib}
\setlength{\itemsep}{-1pt}

\end{document}